%% file: ranlp2025.tex
\documentclass[11pt,a4paper]{article}
\usepackage[hyperref]{ranlp2025}
\usepackage{times}
\usepackage{latexsym}
\usepackage{graphicx}
\usepackage{amsmath}
\usepackage{url}
\usepackage{listings}
\usepackage{booktabs}
\usepackage{xcolor} 
\usepackage{array} 
\usepackage{tabularx}    
\usepackage{multirow} 
\usepackage{algorithm}
\usepackage{algpseudocode}
\usepackage{amsmath}
\usepackage{amssymb}
\usepackage{float}
\usepackage{subcaption}

\newcommand{\norm}[1]{\left\lVert#1\right\rVert}
\definecolor{highlightblue}{RGB}{30, 100, 200}
\definecolor{highlightgreen}{RGB}{50, 140, 90} 

\aclfinalcopy 
\def\aclpaperid{137} 

\title{Am I Blue or Is My Hobby Counting Teardrops? Expression Leakage in Large Language Models as a Symptom of Irrelevancy Disruption}

\author{
\begin{tabular}[t]{c}
\textbf{Berkay K\"opr\"u}$^{1}$ \quad
\textbf{Mehrzad Mashal}$^{2}$ \quad
\textbf{Yigit Gurses}$^{2}$ \quad
\textbf{Akos Kadar}$^{2}$ \quad \\
\textbf{Maximilian Schmitt}$^{1}$ \quad
\textbf{Ditty Mathew}$^{1}$ \quad
\textbf{Felix Burkhardt}$^{1}$ \quad \\
\textbf{Florian Eyben}$^{1}$ \quad
\textbf{Björn W. Schuller}$^{1,3,4}$
\end{tabular}
\\
$^{1}$audEERING GmbH, Gilching, Germany \quad
$^{2}$Agile Robots SE, Munich, Germany \quad \\
$^{3}$Chair of Health Informatics, Technical University of Munich, Germany \quad \\
$^{4}$Group on Language, Audio \& Music, Imperial College London, U.\,K. 
\\ 
\{ \texttt{bkopru, mschmitt, dmathew, fburkhardt, fe, bs\}@audeering.com} \quad \\
\{ \texttt{mehrzad.mashal, yigit.gurses, akos.kadar\}@agile-robots.com} 
}

\begin{document}
\maketitle

\begin{abstract}
Large language models (LLMs) have advanced natural language processing (NLP) skills such as through next-token prediction and self-attention,
but their ability to integrate broad context also makes them prone to incorporating irrelevant information. Prior work has focused on semantic leakage—bias introduced by semantically irrelevant context. In this paper, we introduce \textit{expression leakage}, a novel phenomenon where LLMs systematically generate sentimentally charged expressions that are semantically unrelated to the input context. To analyse the expression leakage, we collect a benchmark dataset along with a scheme to automatically generate a dataset from free-form text from common-crawl. In addition, we propose an automatic evaluation pipeline that correlates well with human judgment, which accelerates the benchmarking by decoupling from the need of annotation for each analysed model. Our experiments show that, as the model scales in the parameter space, the expression leakage reduces within the same LLM family. On the other hand, we 
demonstrate that expression leakage mitigation requires specific care during the model building process, and cannot be mitigated by prompting. In addition, our experiments indicate that, when negative sentiment is injected in the prompt, it disrupts the generation process more than the positive sentiment, causing a higher expression leakage rate.
\end{abstract}

\input{intro}
\input{method}

\input{experiments}
\input{conclusion}

\bibliographystyle{acl_natbib}
\bibliography{ranlp2025}

\end{document}

%% file: intro.tex
\section{Introduction}

Large language models (LLMs) have achieved substantial success by modelling language through next-token prediction. Language model based solutions outperform traditional approaches at the major natural language processing downstream tasks like sentiment analysis \cite{smilga2025scalingsemanticleakageinvestigating}, and named-entity-recognition \cite{kopru23}. LLMs 
owe their success amongst other the self-attention mechanism, introduced in \cite{attentionAllVaswani17} as a core architectural innovation, which supports this formulation and allows effective contextual reasoning.  Self-attention enables models to capture rich contextual dependencies, but it also allows irrelevant or sentimentally charged tokens to influence unrelated generations, contributing to undesired behaviours. Training dynamics, particularly in instruction tuning or reinforcement learning phases, can amplify or suppress such
unintended associations, shaping how strongly sentiment cues propagate through attention pathways.

As the next-token generation LLMs learnt to generate statistically coherent outputs, there is no enforce on factual consistency \cite{huang2025survey}, resulting in hallucinations. In addition to hallucinations, 
systematic and replicable deviations from fairness arises due to the unbalanced nature of data which is free-form text in domain, genre, and cultural representation \cite{navigli2023biases}. 

Beyond social and demographic biases, LLMs also exhibit more subtle generalisation failures, where irrelevant contextual cues influence generation in unintended ways. \citet{shi2023largelanguagemodelseasily} show that LLMs often fail to ignore irrelevant tokens. As a result of irrelevant information, the problem solving capabilities of the LLMs deteriorated. Further,  \citet{gonen2024doeslikingyellowimply} define \textit{Semantic Leakage}, where injecting a semantically unrelated concept biases generation. \citet{smilga2025scalingsemanticleakageinvestigating} further shows that semantic leakage severity increases with model scale. However, most semantic leakage studies focus on concepts and facts being unrelated; none of them explored expressively valenced outputs due to semantically irrelevant context. Moreover, research on \textit{semantic leakage} suffers from a key limitation: it is typically quantified by measuring similarity between discrete lexical entities (e.g., words and sentences). However, such comparisons are ill-defined, as studies on \textit{semantic leakage} presuppose that words and sentences occupy comparable semantic subspaces. 

In this work, we introduce a new phenomenon, \textit{Expression Leakage}, where \textbf{sentimentally} expressive, yet semantically unrelated additions to the prompts cause consistent shifts in LLM outputs. 
Expression leakage and semantic leakage are both forms of unintended generalization, but they differ fundamentally in what is being leaked and how it influences generation. Expression leakage refers to affective or stylistic drift—changes in the emotional tone or sentiment of the output—triggered by sentimentally charged but semantically irrelevant content in the prompt. In contrast, semantic leakage involves the unintended activation of conceptual or factual associations based on semantically unrelated inputs. While semantic leakage alters what is being said, expression leakage alters how it is being said. 

To evaluate how the aforementioned drift in the context affects the generation of LLMs we curated a dataset with 60 samples` and in total 180 prompts. An automatic expression leakage evaluation framework is proposed based on an expression estimator, and via this methodology, it is shown that state-of-the-art LLMs leak expression. Finally, we show that this leakage could be alleviated by prompting especially for instruct models. To summarise our contributions:

\begin{itemize}
    \item Introduce a new phenomenon, \textit{Expression Leakage}, where sentimentally expressive yet semantically unrelated prompt additions cause consistent shifts in LLM outputs.
    \item Curate a controlled benchmark dataset of 60 samples (180 prompt variations) for systematic evaluation of expression leakage.
    \item Develop an automatic data generation scheme producing 200 samples (600 prompt variations) to support scalable evaluation.
    \item Design an automatic evaluation framework that quantifies leakage based on affective shifts at the output level using an expression estimator.
    \item Provide empirical evidence that state-of-the-art LLMs exhibit measurable expression leakage across sentiment conditions.
    \item Demonstrate that leakage can be mitigated through targeted prompting, particularly in instruction-tuned models.
\end{itemize}

\section{Related Work}

\subsection{Semantic Leakage in Language Models}
\citet{gonen2024doeslikingyellowimply} introduced and systematically characterised semantic leakage
effect across 13 LLMs and multiple languages, showing that leakage persists across architectures and appears even in simple, controlled scenarios. Their evaluation strategy relies on embedding-based similarity measures between concepts and generations, identifying subtle forms of concept activation in outputs. \citet{smilga2025scalingsemanticleakageinvestigating} further explore how model scale interacts with semantic leakage, finding that smaller models generally exhibit less leakage, although the relationship is not monotonic, and some intermediate-scale models show localised increases. While these studies establish semantic leakage as a generalisation pathology in LLMs, they focus primarily on factual and lexical associations, without examining affective or stylistic drift.

\begin{table*}[!th]
    \small
    \centering
    \caption{Samples from the HEXL dataset where control and test prompt stem from the curated dataset. Before feeding the prompt to the model, a ``Complete the sentence'' instruction is prepended to all curated prompts.}
    \label{tab::sample_from_human_dataset}
    \begin{tabularx}{\textwidth}{|>{\bfseries}l|X|l|} 
        \hline
        \textbf{Prompt Type} & \textbf{Prompt} & \textbf{Injected Expression} \\ 
        \hline\hline 
        Control & Her passion is & -- \\ 
        \hline
        \multirow{3}{*}{\textbf{Test}} 
            & \textcolor{highlightgreen}{I received a heartfelt compliment from a stranger.} Her passion is & Positive \\ 
            \cline{2-3} 
            & \textcolor{highlightgreen}{I walked down the hallway.} Her passion is & Neutral \\ 
            \cline{2-3}
            & \textcolor{highlightgreen}{I lost my keys on the way here.} Her passion is & Negative \\ 
        \hline
        \hline\hline 
        Control & The music sounded is & -- \\ 
        \hline
        \multirow{3}{*}{\textbf{Test}} 
            & \textcolor{highlightgreen}{I unwrapped an unexpected gift this morning.} The music sounded is & Positive \\ 
            \cline{2-3} 
            & \textcolor{highlightgreen}{I sat on the nearest bench.} The music sounded is & Neutral \\ 
            \cline{2-3}
            & \textcolor{highlightgreen}{I missed the morning bus again.} The music sounded is & Negative \\ 
        \hline
    \end{tabularx}
\end{table*}
\subsection{Sentiment Sensitivity and Affective Misalignment}
Another relevant body of work investigates how LLMs represent and propagate sentiment. \citet{zhang2024sentiment} benchmark LLMs on several sentiment classification datasets, noting that while models perform competitively in binary classification, their robustness degrades in fine-grained or ambiguous contexts. Moreover, \citet{gandhi2025promptsentimentcatalystllm} demonstrate that LLMs often amplify emotional framing embedded in the prompt, even when such framing is irrelevant to the target task. This suggests that LLMs lack the capacity to fully separate semantic content from stylistic or affective signals, particularly when no explicit instruction is given. These findings provide indirect evidence for a form of sentiment-based leakage, but stop short of formalising it as a separate phenomenon or proposing targeted evaluation strategies.

\subsection{Privacy and Information Leakage}
Beyond semantic and sentiment-level shifts, a number of studies address privacy-related information leakage in LLMs. \citet{plant2022privacy} demonstrate that large models can memorise and regurgitate personally identifiable or sensitive content from pretraining data. Their work highlights the risks associated with open-domain generation when model internals encode training data too tightly. \citet{wang2024unlearning} propose machine unlearning techniques to address this concern, showing how forgetting can be selectively induced in certain model layers. Although these works do not address affective leakage directly, they share with our study a focus on unintended generalisation and model behaviour not directly traceable to explicit prompt intent.

\subsection{Cross-Lingual and Stereotype Leakage}
Bias propagation across languages is another form of unintended generalisation studied under the term stereotype leakage. \citet{cao2023stereotypes} explore this in multilingual LLMs, showing that stereotypes embedded in one language can influence outputs in others. This suggests that LLMs encode generalisable latent associations that transcend surface-level linguistic form.  Although stereotype leakage typically targets demographic or social categories, its propagation mechanism is conceptually related to affective leakage across emotional prompts.

\subsection{Positioning Expression Leakage}
Building on these threads, our work proposes \textit{expression leakage} as a novel subclass of semantic leakage—one that manifests not in factual distortions or social biases, but in unintended shifts in affective framing and sentiment intensity. While previous work has identified model sensitivity to emotional cues, there is little systematic evaluation of how irrelevant expressive content in prompts influences model behaviour. Our proposed evaluation framework bridges this gap by quantifying sentiment shifts across matched test and control generations, using an external expression estimator as a proxy for affective alignment. By controlling for prompt structure and target label, we isolate affective leakage and show how it correlates with model size and instruction tuning. This positions expression leakage as both a practical and conceptual extension of semantic leakage—one that captures a previously unquantified vulnerability in large language models.


%% file: method.tex
\section{Methodology}

In this study, we define expression leakage as an extension of \textit{semantic leakage}, and first present a novel dataset generation scheme in which expression leakage rates can be measured for generative models. Then, we describe an automatic benchmarking scheme for expression leakage that decouples the process from the need for a human annotator.

\subsection{Dataset Collection \& Generation}

We follow the definitions by \citet{gonen2024doeslikingyellowimply} and for each sample in the dataset, we come up with a control prompt to observe the state of norm. Then, for each control prompt, we design three test prompts to observe the change of behaviour when a `positive', `neutral', and `negative' sentiment is injected. 

The injected prompts to the control prompt are all in the form of a sentence, and regardless of the desired `expression' type, we use a similar number of tokens per control prompt to remove the effect of context length. The expression of the injected prompt is reviewed by the authors
and the majority vote is taken as the final expression. Following these principles, we collected a dataset we named HEXL (Human-generated EXpression Leakage dataset), $D^H$ 
which contains 60 samples and in total 180 prompts, and an example sample from this curation is provided in Table~\ref{tab::sample_from_human_dataset}. Our human evaluation involved three annotators with backgrounds in linguistics, using majority voting to label expression leakage instances; agreement was above $85\% $ for all categories. Annotators were presented with anonymized control/test generations and asked to judge affective influence blind to the injected sentiment. To scale benchmarking beyond human-curated cases and remove reliance on manual annotation, we formulate an automated data generation scheme for expression leakage using a biased estimator of expression. For each expression class (positive, neutral, negative), we first select the top $m$ sentences with the highest class confidence scores. From this set, we sample $n$ sentences using probabilities proportional to their confidence scores. Neutral sentences are then split into two groups: test candidates and control prompts. Control prompts are further truncated into shorter phrases to serve as flexible suffixes for controlled generation.

\begin{algorithm}[H]
\small
\caption{Sentiment-Controlled Dataset Generation}
\label{alg::sentiment-dataset}
\begin{algorithmic}[1]

\State \textbf{Input:} Sentence dataset $\mathcal{D}$, sentiment model $M$, labels $\mathcal{L}$
\State \textbf{Output:} Labelled prompt dataset $\mathcal{D}^{\text{H}}$

\State Compute sentiment scores for sentences using $M$
\ForAll{$l \in \mathcal{L}$}
    \State Select top-$m$ sentences with highest $p(l)$
    \State Sample $n$ sentences weighted by $p(l)$
\EndFor

\State Split neutral samples into: neutral set and control prompts
\State Truncate control prompts to shorter phrases

\For{$i = 1$ to $k$}
    \State Combine each sentiment sentence with a control prompt
    \State Label each pair with sentiment $l$
\EndFor

\State Save $\mathcal{D^H}$

\end{algorithmic}
\end{algorithm}

In the final stage, each control prompt is combined with sampled sentiment sentences to generate labelled prompts. This pairing enables evaluation or training tasks where the same control input is tested across varying sentiment conditions. The process is summarised in Algorithm~\ref{alg::sentiment-dataset} where AEXL (Automatically Generated Expression Leakage Data Set), $D^A$
is an automatically generated data set.

\subsection{Evaluation Metrics}

To evaluate how much the models leak the injected expression, following \cite{gonen2024doeslikingyellowimply} we define ``mean expression leakage rate" via a biased estimator based on language models. The evaluation scheme includes a preprocessing  on the generated outputs, and leakage decision per generated output. The preprocessing scheme first splits the generated output into sentences, and the first sentence in this split is selected for the evaluation. The selected sentence is cleaned by removing the repetition of the input prompt. The expression score and class of the final cleaned generation is estimated for control and each injected test case. 
Then 

\begin{equation}
\text{EL} = 
\begin{cases}
1, & \text{if } P_{\text{test}}{[l]} > P_\text{ctl}{[l]} \\
0, & \text{otherwise}
\end{cases},
\label{eq::leakage-detection}
\end{equation}

where $P_{\text{test}}, P_{\text{ctl}}\in  \mathbb{R}^3$, where $l \in \{0,1,2\}$ corresponding to the truth value in the injected expression. Then, the leak rate is the average over the whole dataset.

Following \citet{gonen2024doeslikingyellowimply} and \cite{smilga2025scalingsemanticleakageinvestigating}, we compare models via mean-expression leak rate which is the leak-rate calculated  via \ref{eq::leakage-detection} averaged across
dataset:

\begin{equation}
    \mu_{EL,i} = \frac{1}{N_i} \sum_{j=0}^{N_i} EL_j
    \label{eq::mean-exp-leakage-detection},
\end{equation}

where $i \in {D^H, \ D^A}$, and $N_i$ is the number of samples in the corresponding dataset.

To assess whether the test generations exhibit higher expression leakage than their corresponding control generations, we compute the paired difference scores:

$
d_j = P_{\text{test},j} - P_{\text{control},j}.
$

We then apply the one-sided Wilcoxon signed-rank test $W_{\text{EL}}$,
to the difference vector $\mathbf{d} = \{d_1, d_2, \ldots, d_N\}$, testing the alternative hypothesis $H_1: \text{median}(d_j) > 0$. This allows us to determine whether the increase in leakage under the test condition is statistically significant across prompts. We set our significance level to $\alpha=0.001$. 

In addition to the expression leakage, we benchmark the models' performances via semantic leakage on the described datasets. Given a non-linear projector capable of representing properties of a sentence in a vector $\phi: \tau \rightarrow \mathbb{R}^d$, where $d$,
then given a concept $\tau_{concept}$ and two generations $\tau_{ctl}$, $\tau_{test}$: 

\begin{align}
\begin{split}
        sim_{\text{ctl}}  &= \frac{ \phi(\tau_{\text{concept}}) \cdot \phi( \tau_{\text{ctl}} )}{\norm{\phi( \tau_{\text{concept}} )} \cdot \norm{\phi( \tau_{\text{ctl}} )}}, \\
        sim_{\text{test}}  &= \frac{ \phi(\tau_{\text{concept}}) \cdot \phi( \tau_{\text{test}} )}{\norm{\phi( \tau_{\text{concept}} )} \cdot \norm{\phi( \tau_{\text{test}} )}}.
\end{split}
\end{align}

Then, semantic leakage per prompt is decided using the following heuristic: 

\begin{equation}
\text{L} = 
\begin{cases}
1 & \text{if}\ sim_{\text{test}} >  sim_{\text{ctl}} \\
0 & \text{if}\ sim_{\text{test}} <  sim_{\text{ctl}} \\
0.5 & \text{otherwise}
\end{cases}.
\label{eq::sem-leakage-detection}
\end{equation}

As in Eq.~\ref{eq::mean-exp-leakage-detection}, the semantic leakage rate of the model is defined as the average over the dataset with $\mu_{L,i}$ where $i \in {D^H, \ D^A}$.


\subsection{Models}

We focus on evaluating two types of consumer-grade models ranging between 1.5B–8B parameters: (1) general-purpose models and (2) instruction-tuned models.
Taking into account these criteria, the Qwen2.5 \cite{qwen2025qwen25technicalreport}, and Llama3.2 \cite{grattafiori2024llama3herdmodels} model families are selected for the evaluation expression leakage.

%% file: experiments.tex
\section{Experiments}

State-of-the-art LLMs are evaluated on the proposed datasets and their performances are compared in terms of expression and semantic leakage rates. In this section, we first introduce our experimentation setup, then, the experimental results are presented and discussed.
\begin{figure*}[ht]
    \centering
    \begin{subfigure}[b]{0.48\linewidth}
        \centering
        \includegraphics[width=0.8\linewidth]{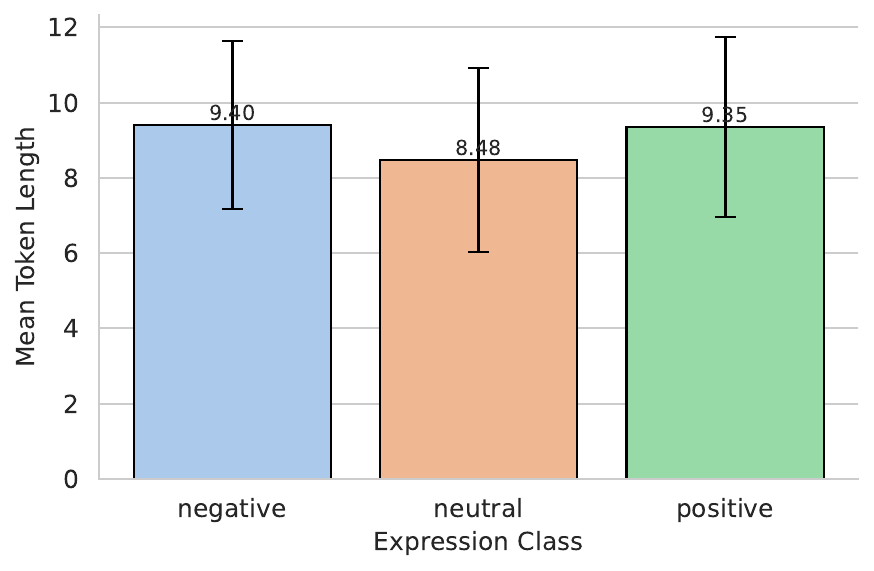}
        \caption{Mean-injected token lengths per expression class on the human generated expression leakage dataset (HEXL)}
        \label{fig:token_len}
    \end{subfigure}
    \hfill
    \begin{subfigure}[b]{0.48\linewidth}
        \centering
        \includegraphics[width=0.8\linewidth]{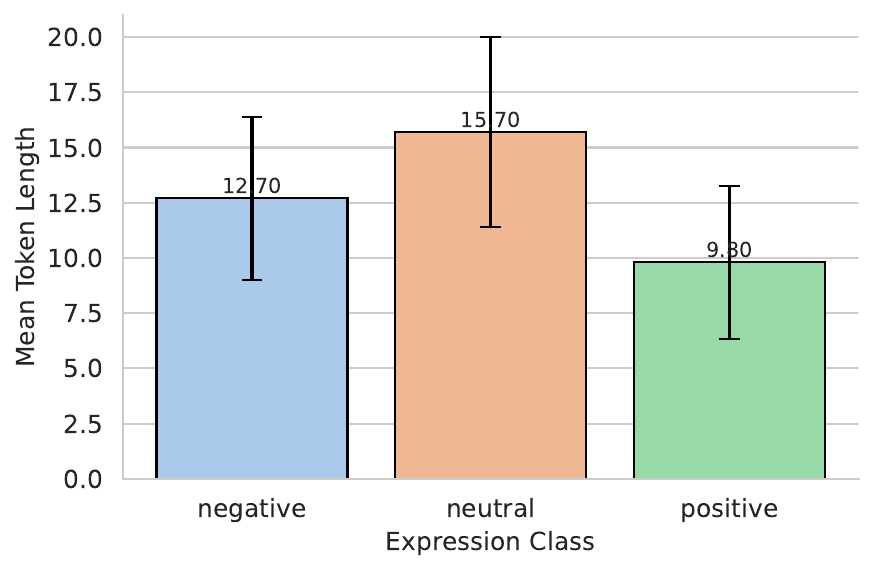}
        \caption{Mean-injected token lengths per expression class on the automatically generated expression leakage dataset (AEXL)}
        \label{fig:token_len_b}
    \end{subfigure}
    \caption{Comparison of injected prompt characteristics across sentiment dimensions}
    \label{fig:token_len_automatic}
\end{figure*}

\subsection{Experimental Setup}

{\scriptsize
\begin{table*}[!ht]
    \small
    \centering
    \caption{Sample generations from the Qwen and Llama families on the HEXL dataset. The text highlighted with green corresponds to injected prompts, while blue corresponds to the model's generation.}
    \label{tab::sample_generation_from_human_dataset}
    \begin{tabularx}{\textwidth}{|
        >{\raggedright\arraybackslash}p{2cm}|
        >{\raggedright\arraybackslash}p{1.5cm}|
        >{\scriptsize\raggedright\arraybackslash}X|
        >{\raggedright\arraybackslash}p{2.2cm}|}
        \hline
        \textbf{Model} & \textbf{Prompt Type} & \normalsize\textbf{Prompt}  & \textbf{Injected Expression} \\ 
        \hline\hline
        \multirow{4}{=}{Llama-3.2-1B-Instruct} 
            & Control & The painting reveals \textcolor{highlightblue}{a hidden world of ancient civilizations and mystical creatures that had been lost to time.} & -- \\ 
            \cline{2-4}
            & \multirow{3}{*}{Test} 
            & \textcolor{highlightgreen}{She found a long-lost letter in her attic.} The painting reveals \textcolor{highlightblue}{a hidden message that had been encoded on it for decades.} & Positive \\ 
            \cline{3-4}
            &  
            & \textcolor{highlightgreen}{She checked the time on her watch.} The painting reveals \textcolor{highlightblue}{a faint message that read ``I was here".} & Neutral \\ 
            \cline{3-4}
            &  
            & \textcolor{highlightgreen}{She received an angry phone call.} The painting reveals \textcolor{highlightblue}{a dark secret about her husband's past.} & Negative \\ 
        \hline\hline 
        \multirow{4}{=}{Qwen2.5-1.5B-Instruct} 
            & Control & The painting reveals \textcolor{highlightblue}{a hidden world of ancient civilizations and mystical creatures that had been lost to time.} & -- \\ 
            \cline{2-4}
            & \multirow{3}{*}{Test} 
            & \textcolor{highlightgreen}{She found a long-lost letter in her attic.} The painting reveals \textcolor{highlightblue}{she has always loved this place.} & Positive \\ 
            \cline{3-4}
            &  
            & \textcolor{highlightgreen}{She checked the time on her watch.} The painting reveals \textcolor{highlightblue}{a hidden message within it.} & Neutral \\ 
            \cline{3-4}
            &  
            & \textcolor{highlightgreen}{She received an angry phone call.} The painting reveals \textcolor{highlightblue}{she had been burglarized.} & Negative \\ 
        \hline
    \end{tabularx}
\end{table*}
}

Experiments were conducted using NVIDIA H100 GPUs (80\,GB) on a single GPU with PyTorch 2.5.1 and the Hugging Face Transformer v4.49.0. In addition, for the generation of AEXL, we used a dataset\footnote{https://huggingface.co/datasets/agentlans/high-quality-english-sentences} hosted at HuggingFace which is a filtered version of C4 \cite{raffel2020exploring} and FineWeb \cite{penedo2024the}. While calculating expression leakage ($\mu_{\text{EL}}$), we consider only the injected cases of positive and negative sentiment, as the neutral injection condition is semantically aligned with the control and thus directly captured by the semantic leakage metric ($\mu_l$).
Next, we present the models and inference configurations used in the experiments. 

\subsubsection{Models}
\begin{table*}[h]
    \centering
    \begin{tabular}{lcccccc}
    \toprule
    \textbf{Model} & \multicolumn{3}{c}{\textbf{HEXL }} & \multicolumn{3}{c}{\textbf{AEXL}} \\
                  & $\mu_l$ & $\mu_{\text{EL}}$ & $W_{\text{EL}}$ & $\mu_l $ & $\mu_{\text{EL}}$ & $W_\text{EL}$\\
    \midrule
    Llama-3.2-1B          & 0.83              & 0.73        & 9.39e-9              & 0.75        & 0.66 & 1.34e-15 \\
    Llama-3.2-3B          & 0.82              & 0.75        & 1.41e-9              & 0.67        & 0.64 & 3.20e-20 \\
    Llama-3.1-8B          & 0.69              & 0.74        & 6.50e-9              & 0.65        & 0.61 & 4.2e-32 \\
    Llama-3.2-1B-Instruct & 0.86              & 0.73        & 7.42e-6              & 0.84       & 0.80 & 5.47e-39 \\
    Llama-3.2-3B-Instruct & 0.79              & 0.67        & 2.82e-7              & 0.79        & 0.72 & 1.31e-31 \\
    Llama-3.1-8B-Instruct & 0.78              & 0.65        & 1.05e-5              & 0.78        & 0.66 & 1.53e-5 \\
    \hline \hline
    Qwen2.5-0.5B          & 0.91              & 0.74        & 2.50e-11              & 0.85        & 0.76 & 1.69e-30 \\
    Qwen2.5-1.5B          & 0.79              & 0.70        & 3.67e-7               & 0.82        & 0.73 & 3.33e-35 \\

    Qwen2.5-3B            & 0.84              & 0.68       & 3.68e-9              & 0.79        & 0.71 & 2.45e-25 \\
    Qwen2.5-0.5B-Instruct & 0.87              & 0.77        & 1.99e-11              & 0.91        & 0.82 & 2.37e-44 \\
    Qwen2.5-1.5B-Instruct & 0.78              & 0.73        & 1.39e-8              & 0.87        & 0.75 & 3.20e-35 \\
    Qwen2.5-3B-Instruct   & 0.79              & 0.67        & 6.47e-8              & 0.79        & 0.69 & 1.06e-33 \\
    \bottomrule
    \end{tabular}
    \caption{Mean expression and sentiment leakage rates across models and datasets.}
    \label{tab::leakage_summary}
\end{table*}

Focusing on consumer-grade LLMs and their instruction-following variants, we selected the Qwen2.5 \cite{qwen2025qwen25technicalreport} and Llama 3 \cite{grattafiori2024llama3herdmodels} model families for our evaluation of expression leakage


For the expression leakage evaluation, we make use of the sentiment classifier proposed by \cite{barbieri-etal-2020-tweeteval}, which is a RoBERTa \cite{liu2019robertarobustlyoptimizedbert} based model trained on Twitter data. We especially select a model that is trained on a Twitter dataset considering LLMs might produce emojis and emoticons, and models trained on Twitter data handle these information better than structured-text data. We use the \textit{cardiffnlp/twitter-roberta-base-sentiment} model.\footnote{https://huggingface.co/cardiffnlp/twitter-roberta-base-sentiment} For the semantic leakage evaluation, we use SentenceBERT \cite{reimers2019sentencebertsentenceembeddingsusing} to project sentences into vectors where similarity can be calculated.
\subsubsection{Inference Configurations}

For the evaluation of semantic leakage, we generate model outputs using a controlled decoding configuration designed to introduce lexical variability while preserving semantic coherence. Specifically, we decide for nucleus sampling with a top-$p$ value of 0.9 and a top-$k$ cutoff of 50 to encourage diverse, but high-probability continuations. Each prompt is decoded by sampling with a repetition penalty of 1.1 to reduce verbatim regeneration, and a maximum generation length of 128 tokens is enforced to ensure bounded output. To enable statistical robustness in leakage estimation, 10 generations are produced per prompt, allowing for instance-level aggregation across output samples.

\subsection{Injected Context Length Analysis}

In \cite{CHENLO2014275} it is shown that the sentence length induces a bias on the sentiment, and in addition to sentiment bias, the longer contexts also might introduce additional semantical distractions due to  the self-attention mechanism. The human generated dataset process does not have a restriction on the prompt length. Considering context length bias on sentiment and semantics,  we analysed the length of injected context per expression class. Figure~\ref{fig:token_len} depicts that the injected contexts are in similar length in terms of number of tokens, when measured with the GPT-2 tokenizer \cite{radford2019languageGPT2} across the expression categories where for the negative class the mean is $9.40$,  for the neutral class it is $8.48$, and for the positive class it is $9.35$. On the right hand side, Figure~\ref{fig:token_len_automatic} depicts the token lengths for the AEXL dataset, where at each sentiment category, the prompts are longer. For the negative class, the mean token length is $12.70$, for the neutral class it is $15.70$, and for the positive class it is $9.80$. The difference in the neutral class suggests that during the curation of the HEXL dataset, curators instinctively generate a uniform distribution, and in the free-form webscale data, neutral sentences are longer than the rest of the other categories.

\subsection{Automatic Expression Leakage Rate and Semantic Leakage Rate} \label{sec::exp-leakage rate}

Table~\ref{tab::leakage_summary} presents the mean-expression leakage rate ($\mu_{\text{EL}}$) along with the mean-semantic leakage rate calculated over the HEXL and AEXL datasets. For
all models and datasets, the $\mu_{\text{EL}}$ exceeds
$0.5$, meaning that test generations show higher expression similarity to the corresponding concepts in comparison with control generations. This implies that a leakage-rate $\mu_{\text{EL}}=0.5$ corresponds to random-similarity and no leakage. Observing the $\mu_{\text{EL}}$ performances, the Qwen family exhibits a lower leakage rate compared to the Llama family in the comparable model sizes. Comparing the effect of the parameter size on the $\mu_{\text{EL}}$, it is observed that the model size positively effects the expression leakage, where more capable models leak less expression. The Llama-3.2 family reduced the leakage by 16 points while scaling by 8 in their parameter space. This finding contradicts with the negative correlation between semantic leakage and scale of the models depicted in \cite{smilga2025scalingsemanticleakageinvestigating}.

Another contradiction of expression leakage against expression leakage 
is observed: instruct models are more capable to disregard the irrelevant expression: comparing Llama-3.2-3B-Instruct to the base model Llama-3.2-3B, there is an 8 percent reduction in the leakage rate. This could be attributed to the fact that instruct models are probably more capable of understanding the expression than the base models are. 
\begin{figure}[!h]
    \centering
    \includegraphics[width=\linewidth]{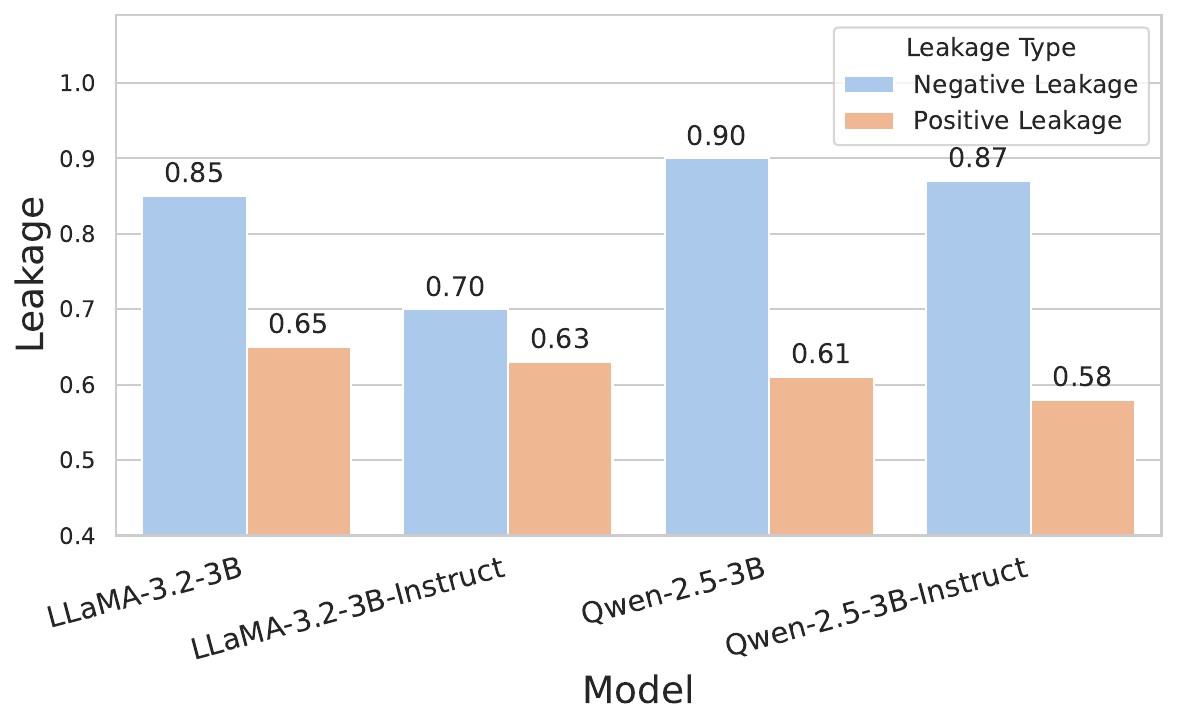}
    \caption{Comparison of same size LLMs for their expression leakage rate per injected expression}
    \label{fig::positive-vs-neg-leakage}
\end{figure}
We used the Wilcoxon signed rank test to assess the statistical significance of leakage differences between models. It is well-suited for our paired, non-normally distributed data, providing a robust alternative to parametric tests. This allows us to detect consistent shifts in leakage behaviour and validate the impact of model variants or training strategies. 

Figure~\ref{fig::positive-vs-neg-leakage} depicts the comparison of leakage performance across same size models per injected expression. We observe a consistent pattern where negative sentiment injections result in higher leakage rates than positive ones, particularly in base models. This asymmetry may stem from the model's heightened sensitivity to negatively framed expressions, which often carry stronger affective or directive cues that align more easily with underlying training data patterns. Additionally, negative prompts may trigger more deterministic or safety-related completions, increasing the likelihood of unintended leakage. In contrast, positive expressions tend to be more diffuse or open-ended, making it harder for the model to converge on specific, memorised outputs. This suggests that the semantic framing of prompts significantly influences the model’s response bias and its susceptibility to information leakage.
\begin{table*}[!th]
    \centering
    \begin{tabular}{lcccccc}
    \toprule
    \textbf{Model} & \multicolumn{2}{c}{\textbf{HEXL }} & \multicolumn{2}{c}{\textbf{AEXL}} \\
                  & $\mu_{\text{EL}}$ & $W_{\text{EL}}$ &  $\mu_{\text{EL}}$ & $W_\text{EL}$\\
    \midrule

    Llama-3.2-1B-Instruct & 0.69              & 1.04e-7        & 0.78              & 3.67e-37         \\
    Llama-3.2-3B-Instruct & 0.68              & 2.90-6        & 0.71              & 1.48e-21         \\
    Llama-3.1-8B-Instruct & 0.60              & 2.23e-7        & 0.70              & 1.63e-27         \\

    Qwen2.5-0.5B-Instruct & 0.77              & 6.33e-12
        & 0.85              & 8.94e-51         \\
    Qwen2.5-1.5B-Instruct & 0.73              & 1.33e-8        & 0.86              & 5.76e-52         \\
    Qwen2.5-3B-Instruct   & 0.71              & 2.05e-6        & 0.70              & 2.83e-31         \\

    \bottomrule
    \end{tabular}
    \caption{Mean expression leakage rates across models and datasets with additional instruction to disregard irrelevant information.}
    \label{tab::disregard_leakage_summary}
\end{table*}
On the AEXL benchmark, expression leakage remains prominent across most models, with $\mu_{\text{EL}}$ values closely aligned with those observed on HEXL. Instruction-tuned variants generally exhibit higher leakage than their base counterparts. For instance, Qwen2.5-0.5B-Instruct reaches a leakage rate of $\mu_{\text{EL}} = 0.82$, compared to $0.76$ in the untuned Qwen2.5-0.5B model. This trend suggests that while instruction tuning improves usability, it may also increase sensitivity to sentimentally expressive, but irrelevant prompt cues. Wilcoxon test scores ($W_{\text{EL}} < 0.001$) confirm that these differences are statistically significant across models,
reinforcing that model tuning and prompt design play a critical role in mitigating or amplifying expression leakage.

Semantic leakage results reveal that within the Llama base models, leakage decreases with scale—dropping from 0.83 to 0.69 on HEXL and from 0.75 to 0.65 on AEXL. However, instruction-tuned variants do not consistently reduce leakage: for instance, Llama-3.2-1B-Instruct shows higher leakage than its base counterpart. In the Qwen family, leakage remains high across all scales, and instruction tuning often amplifies it, as seen with Qwen2.5-0.5B-Instruct reaching 0.91 on AEXL. These findings partly contrast with \cite{smilga2025scalingsemanticleakageinvestigating}, who observed that smaller models can leak less, attributing higher leakage in larger models to increased associative capacity. Notably, trends in semantic and expression leakage are broadly aligned across HEXL and AEXL: models that show high semantic leakage tend to exhibit high expression leakage as well, suggesting shared underlying generalisation behaviours across both forms of leakage.

\subsection{Alleviating Expression Leakage via Prompting}

As instruct models finetuned over base-models tend  
to comply directions and improve capabilities in downstream tasks, we study whether explicitly adding an instruction to disregard irrelevant information alleviates the leakage rates of these models. For this study, we prepend following instruction to each prompt: \textit{Ignore any irrelevant information in user prompt that is not relevant to the request}, and re-run our experiments on expression leakage from Section~\ref{sec::exp-leakage rate}.

Table~\ref{tab::disregard_leakage_summary} depicts the expression leakage rates with induced instruction to \textit{disregard irrelevant information}. Comparing performances of Llama-Instruct family in Table~\ref{tab::leakage_summary} and Table~\ref{tab::disregard_leakage_summary}, it is observed that there is no consistent change amongst the models $\mu_{\text{EL}}$. These results suggest that prompting is not enough, and specific mitigation techniques must be applied to alleviate expression leakage rate.


%% file: conclusion.tex
\section{Conclusion}
In this work, we introduced \textit{expression leakage} as a novel form of unintended generalisation in large language models (LLMs), where sentimentally expressive, but semantically irrelevant prompt content induces systematic shifts in model outputs. To evaluate this phenomenon, we curated a controlled benchmark dataset and proposed an automatic expression leakage analysis framework based on a biased expression estimator, which correlates well with human judgment and enables scalable benchmarking without manual annotation. Through extensive experiments, we demonstrated that expression leakage is prevalent across models and that its severity diminishes with increased model scale within the same LLM family, suggesting a connection to capacity and internal control mechanisms. However, we also showed that leakage cannot be mitigated reliably through prompting alone, and that instruction tuning, while partially effective, is not sufficient to suppress affective drift when irrelevant sentiment is introduced. Notably, our results indicate that negative sentiment causes more disruptive leakage than positive sentiment, pointing to an asymmetry in how emotional context influences generation. Expression leakage complements existing notions of semantic and factual leakage by focusing on affective modulation, offering a new lens to evaluate the robustness and controllability of LLMs. We hope this work motivates further research on fine-grained bias detection and the development of training and inference strategies that safeguard against unintended affective influence.